\documentclass{article}

\usepackage{microtype}
\usepackage{graphicx}
\usepackage{subfigure}
\usepackage{booktabs} 
\usepackage{xcolor}
\usepackage{amsmath}
\usepackage{amssymb}
\usepackage{amsthm}
\usepackage{bm}
\usepackage{graphicx}
\usepackage{wrapfig}
\usepackage{diagbox}
\usepackage{multirow}
\usepackage{enumitem}

\usepackage{hyperref}

\usepackage[accepted]{icml2021}

\icmltitlerunning{Keyframe-Focused Visual Imitation Learning}

\newtheoremstyle{questionstyle}
  {\topsep}   
  {0}         
  {\itshape}  
  {0pt}       
  {\bfseries} 
  {.}         
  {5pt plus 1pt minus 1pt} 
  {}          
\theoremstyle{questionstyle}\newtheorem{question}{Question}

\begin{document}

\twocolumn[

\icmltitle{Keyframe-Focused Visual Imitation Learning}

\icmlsetsymbol{equal}{*}

\begin{icmlauthorlist}
\icmlauthor{Chuan Wen}{equal,iiis}
\icmlauthor{Jierui Lin}{equal,austin}
\icmlauthor{Jianing Qian}{upenn}
\icmlauthor{Yang Gao}{iiis,qizhi}
\icmlauthor{Dinesh Jayaraman}{upenn}
\end{icmlauthorlist}

\icmlaffiliation{iiis}{Institute for Interdisciplinary Information Sciences, Tsinghua University}
\icmlaffiliation{austin}{University of Texas at Austin}
\icmlaffiliation{upenn}{University of Pennsylvania}
\icmlaffiliation{qizhi}{Shanghai Qi Zhi Institute}

\icmlcorrespondingauthor{Dinesh Jayaraman}{dineshj@seas.upenn.edu}

\icmlkeywords{Machine Learning, ICML}

\vskip 0.3in
]

\printAffiliationsAndNotice{\icmlEqualContribution} 

\begin{abstract}
Imitation learning trains control policies by mimicking pre-recorded expert demonstrations. In partially observable settings, imitation policies must rely on observation histories, but many seemingly paradoxical results show better performance for policies that only access the most recent observation. Recent solutions ranging from causal graph learning to deep information bottlenecks have shown promising results, but failed to scale to realistic settings such as visual imitation. We propose a solution that outperforms these prior approaches by upweighting demonstration keyframes corresponding to expert action changepoints. This simple approach easily scales to complex visual imitation settings. Our experimental results demonstrate consistent performance improvements over all baselines on image-based Gym MuJoCo continuous control tasks. Finally, on the CARLA photorealistic vision-based urban driving simulator, we resolve a long-standing issue in behavioral cloning for driving by demonstrating effective imitation from observation histories. Supplementary materials and code at: \url{https://tinyurl.com/imitation-keyframes}.
\end{abstract}

\section{Introduction}

Learning controllers for complex, unmodeled agents and environments is a challenging problem. 
For tasks where at least one ``expert'' controller exists, such as a human driver for autonomous driving, imitation learning offers a simple, powerful family of solutions that exploit demonstrations provided by this expert to bootstrap control policy learning. Many imitation approaches employ a straightforward ``behavioral cloning'' (BC) strategy, to train policies completely ``offline'', i.e., with no environmental interaction, by simply mapping expert observations to expert actions on the demonstration data. While BC has well-documented distributional shift issues due to compounding imitation errors when executed in the environment, several effective approaches have been proposed to address them, and BC remains widely used in practice~\cite{pomerleau1989alvinn, schaal1999imitation, muller2006off, mulling2013learning, DBLP:journals/corr/BojarskiTDFFGJM16, giusti2015machine}.

\begin{figure*}[htb]
    \centering
    \includegraphics[width=0.8\textwidth]{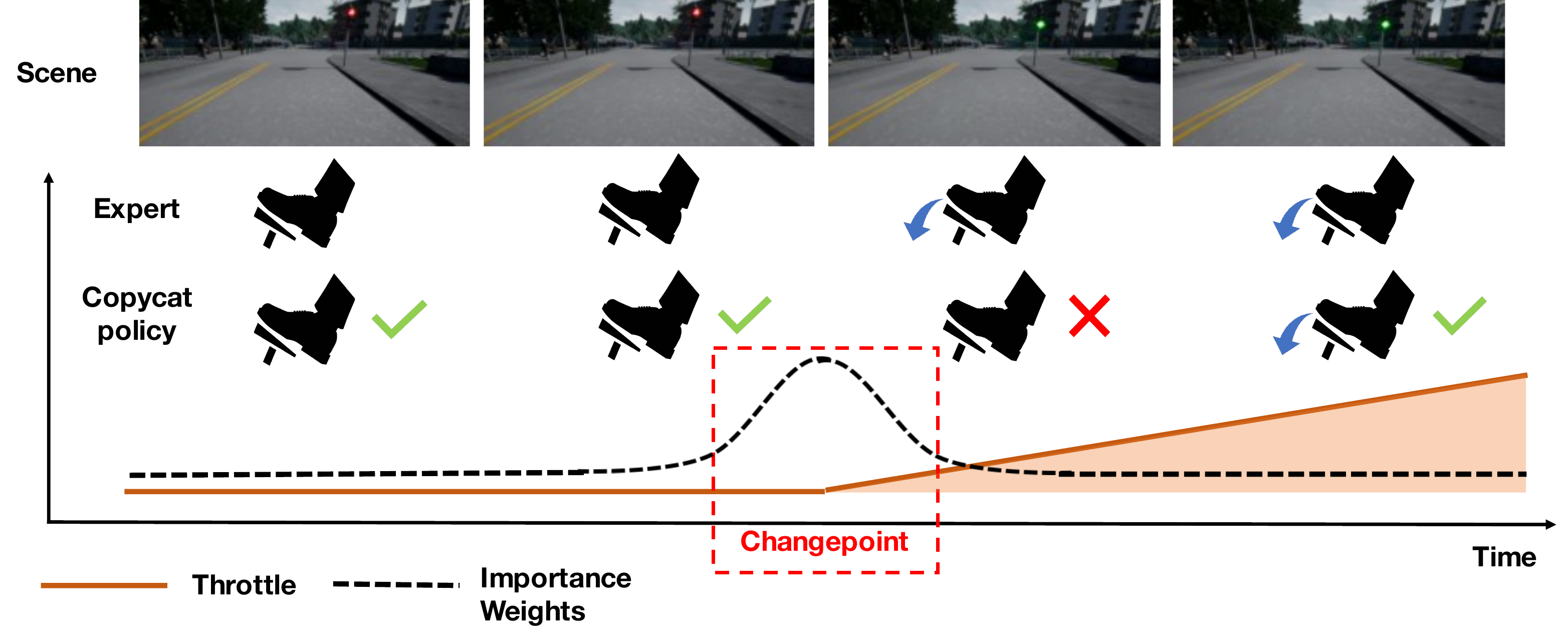}
    \caption{An instance of the copycat issue in the CARLA autonomous driving simulator. Views from the expert data show the policy waiting at a red light and then accelerating (throttle) when it turns green. A simple copycat policy is mostly correct but makes a mistake at this critical keyframe. We define a notion of changepoints to detect such keyframes and upweight them during behavior cloning.}
    \label{fig: copycat}
\end{figure*}

We focus on the open problem of effectively extending BC to realistic partially observed settings such as driving, where the agent cannot observe all task-relevant information instantaneously. This is commonly resolved in other controller design paradigms by integrating historical information in the control policy, but this has proven challenging in BC. For over 15 years now, researchers have reported seemingly paradoxical results that show performance drops in some POMDP settings from allowing BC agents to access history information, compared to when they are restricted to instantaneous observations alone~\citep{muller2006off,bansal2018chauffeurnet,NIPS2019_9343,wang2019monocular}. 
Recently, \citet{chuan2020fighting} coined the phrase ``copycat problem'' to describe the issue, and show that the problem is wider still: even when history information does improve BC performance, the learned policies often perform suboptimally and have room to improve if the copycat problem is correctly addressed.

Figure~\ref{fig: copycat} shows a snippet of an expert driving demonstration from the imitation dataset CARLA100~\cite{Dosovitskiy17,codevilla2019exploring}, where a car waiting at a red traffic light starts to move when the light turns green. We can see that the expert's action  $a_{t}$ is identical to its previous action $a_{t-1}$, except at one moment when the light turns green (figure shows throttle). Thus, a ``copycat'' policy that repeated the previous action without paying any attention to the images would only commit one imitation error on the expert data. Upon execution in the environment however, since no expert would be available, it would repeat its own previous action at each step, and never move at all!
Indeed, \citet{codevilla2019exploring} report this special case of the copycat problem as the ``inertia'' problem.

We study the reasons for the copycat problem and identify one key reason that can be algorithmically addressed: when expert actions are highly temporally correlated, the demonstration dataset has a very tiny fraction of important ``changepoint'' samples that typically correspond to the expert responding to some external change in the observations, such as when the traffic light turns green. We then propose a simple, well-motivated metric to automatically identify these underrepresented changepoint samples in the demonstrations, and propose to upweight them in the behavioral cloning objective function for policy learning. 

We evaluate our method across four varied simulated environments, ranging from robotic control from clean images, to photorealistic urban driving environments. Our experimental results validate that our method offers the most effective and scalable solution yet for tackling the copycat problem, while also being very simple to implement.

\section{Related Work}\label{sec: related-work}

\textbf{Imitation Learning.}
Imitation learning is a powerful policy learning method that can learn complex decision behaviors from expert demonstrations~\cite{widrow1964pattern,osa2018algorithmic, argall2009survey}. In this paper, we focus on the widely used behavioral cloning paradigm of imitation, which directly regresses from observations to expert actions~\cite{pomerleau1989alvinn,schaal1999imitation,muller2006off,mulling2013learning,DBLP:journals/corr/BojarskiTDFFGJM16,giusti2015machine}. Like other imitation approaches, behavioral cloning must contend with distribution shift: small errors between imitator and expert policies accumulate over time leading the imitator into unfamiliar states~\cite{Ross2011}.
It is possible to resolve this through environmental interactions~\cite{NIPS2016_6391, Brantley2020} or queryable experts~\cite{Ross2011,Sun2017,Laskey2017,sun2018truncated}. Our focus is on a specific well-documented problem arising due to distributional shift in partially observed imitation settings, recently coined the ``copycat problem''~\cite{chuan2020fighting}. We discuss work specifically related to the copycat problem in detail in Sec~\ref{sec: introducing-copycat}, after setting the context.

\textbf{Importance Weighting To Tackle Data Imbalance.} 
Sample reweighting / resampling, a well-known technique in machine learning and statistics, has recently been shown to remain very effective at tackling long-tail problems arising from data imbalances in machine learning~\cite{cui2019class,cao2019learning,kang2019decoupling,zhou2020bbn}. \citet{wang2018exponentially} assume access to environmental rewards, and reweight training samples for imitation based on their corresponding value functions. 
While these approaches rely on ``labels'' such as category annotations or environmental rewards, we instead discover an unlabeled group of ``changepoint'' keyframes in imitation learning datasets. By identifying the scarcity of such frames as a data imbalance that causes copycat problems, we are able to propose a simple and surprisingly effective sample reweighting-based technique to alleviate them.

\textbf{Shortcut Learning.}
With the increasingly widespread use of machine learning, researchers have begun to pay attention to various intriguing errors and quirks, particularly with deep neural networks (DNNs). DNN image classifiers often classify images based on irrelevant backgrounds rather than foregrounds~\cite{beery2018recognition} and object textures rather than shapes~\cite{geirhos2018imagenet}. 
\citet{geirhos2020shortcut} recently surveyed several such phenomena, identifying them as instances of ``shortcuts'': models that are \emph{easy} to learn, perform well on the data they were trained on, but then fail to generalize to the real world. We view the copycat problem as another instance of the shortcut learning problem, identify conditions that lead to its emergence in imitation learning, and make progress towards alleviating it.

\section{Preliminaries}

We are interested in learning control policies in settings that can be modeled as partially observed Markov decision processes (POMDP). In POMDPs, the environment at time $t$ provides to the agent a reward $r_t$, and an observation $o_t$ which only partially represents its true state. To account for this missing state information, it is common practice~\cite{murphy2000survey,schulman2017proximal} to augment the current observation $o_t$ with the last $H$ observations to form the ``observation history'' $\tilde{o}_t = [o_{t-H}, \cdots, o_t]$. Optimal controllers that maximize the sum of environmental rewards\footnote{ignoring discount factors for simplicity} $R=\sum_t r_t$ must rely on this observation history $\tilde{o}_t$ rather than solely on $o_t$.

\subsection{Behavioral Cloning}

While the above point about observation histories also holds for control policies synthesized through reinforcement learning or other approaches, we are interested in policies trained via imitation learning. In particular, we focus on the widely used behavioral cloning (BC) paradigm, which reduces imitation to simple supervised learning to mimic expert actions. Specifically, an expert policy $\pi_e$, such as a human demonstrator, generates training demonstrations $\mathcal{D}=\{ (o_{t}, a_{t}) \}_{t=1}^{N}$. The goal of BC is to train a parameterized policy $\pi_{\theta}(\tilde{o}_{t}) = \hat{a}_{t}$ that estimates the expert's action at time $t$.
To do this, BC methods typically minimize the following mean squared error (MSE) loss on $\mathcal{D}$:
\begin{align}
    \mathop{\arg\min}_{\theta} \text{MSE}_\mathcal{D}(\theta) &=
    \frac{1}{N}\sum_{t=0}^N(\pi_\theta(\tilde{o}_t)-a_{t})^{2}.
    \label{eq:bc_loss}
\end{align}

\subsection{The ``Copycat'' Problem in Behavioral Cloning}\label{sec: introducing-copycat}

As mentioned above, optimal controllers typically require historical information to account for partial observability. Therefore, we would expect BC policies with access to the observation history $\tilde{o}_t$ (``BC-OH'') to perform better than those that map a single observation $o_t$ to $a_t$ (``BC-SO''). Yet, in practice, many prior works~\cite{muller2006off,bansal2018chauffeurnet,NIPS2019_9343,chuan2020fighting,codevilla2019exploring,wang2019monocular} report that BC-OH performs poorly compared to BC-SO. In particular, BC-OH produces better (lower) values of the BC loss in Eq~\eqref{eq:bc_loss} on both training and validation data  from expert demonstrations, but performs poorly when actually executed in the environment. In recent attempts to deal with this issue, it has variously been identified as the ``copycat'' problem~\cite{chuan2020fighting}, the ``inertia'' problem~\cite{codevilla2019exploring}, and ``causal confusion''~\cite{NIPS2019_9343}: an imitator exploits the strong temporal correlation of expert actions to learn policies that predict $a_t$ purely as a function of previous actions $a_{t-1}, a_{t-2}, \cdots $. \citet{chuan2020fighting} make two important observations that widen the scope of the copycat problem: (1) Even when history information does improve BC performance as we would expect, the learned policies often perform suboptimally and have room to improve if the copycat problem is correctly addressed. (2) Even when past actions are not explicitly available as input, the imitator commonly learns to recover them from the observation history $\tilde{o}_t$ and manifest the copycat problem.

\section{Method}
We now analyze the copycat problem and identify its key causes. Motivated by this analysis, we then propose a simple approach that aims to resolve the problem by reweighting training data samples based on the temporal characteristics of expert action sequences.

\subsection{What Causes the Copycat Problem?}\label{sec: motivation}

We argue that the copycat problem arises from (\textbf{A}) strong temporal correlation among expert actions, (\textbf{B}) misalignment between environmental reward and the imitation objective, and (\textbf{C}) the difficulty of learning truly optimal policies that fit the expert data. First, temporal correlation makes it possible for a ``copycat policy" $\psi(a_{t-1}, a_{t-2}, .... )$ that relies purely on expert action histories to produce low MSE for predicting expert actions $a_t$ on expert demonstrations (training as well as held-out data) without accessing environmental observations at all. Second, the well-documented distributional shift problem in imitation~\cite{Ross2011}, compounded by misalignment between the MSE objective and the true environment reward $R$, means that $\psi(\cdot)$ yields low rewards upon environmental execution. And finally, it is difficult to learn a ``good'' policy that correctly identifies and relies on the causes of expert actions among the observations. This means that the copycat policy offers an excellent ``shortcut''~\cite{geirhos2020shortcut} to the BC learner. We expand further on these intuitions below. 

Suppose we train an optimal copycat policy $\psi^*(\cdot)$ on the training dataset through behavioral cloning as:
\begin{equation}
    \psi^* = \mathop{\arg\min}_{\psi} \frac{1}{N} \sum_{t=0}^N(\psi (a_{t-1}, a_{t-2}, \cdots) - a_{t})^{2}.
    \label{eq:copycat_loss}
\end{equation}
Suppose further that the expert data has a fraction $\epsilon_{CP}$ of ``changepoint" frames for which $a_t$ is not predicted well by the optimal copycat policy $\psi^*(\cdot)$. For convenience, we will assume these samples all suffer from uniform copycat error equal to $1$, so that the training MSE of $\psi^*(\cdot)$ is the changepoint fraction $\epsilon_{CP}$. Low $\epsilon_{CP}$ corresponds to low-MSE copycats. This relates to \textbf{A} above.

Next, we turn our attention to the reward-optimal policy parameters $\theta^{R*}$ corresponding to the policy within the model class, that yields the highest environmental reward $R$. Observe that, in general,  $\theta^{R^{*}}$ does not fit the expert data perfectly. In other words, it produces a non-zero training error $\text{MSE}_\mathcal{D}(\theta^{R*})>0$. This happens due to the misalignment issue (\textbf{B} above), and optimization difficulties, model class mismatch or noisy demonstrations (\textbf{C} above).

When we synthesize the above observations, a clear-cut domain for the copycat problem begins to emerge. 
BC learners will always prefer the copycat solution $\psi^*$ over the reward-optimal parameters if:
\begin{equation}
    \text{MSE}_\mathcal{D}(\theta^{R*})>\epsilon_{CP}, \label{eq:copycat_condition}
\end{equation}
or in other words, the BC training loss is lower for the copycat $\psi$ than it is for $\pi_{\theta^{R*}}$.\footnote{Since $\psi$ is typically a very simple function, we implicitly make the assumption that the learning algorithm can easily find parameters $\theta$ such that $\pi_\theta(\cdot)=\psi^*(\cdot)$.} 
Note that we operate in data-rich settings without overfitting, so that all the above statements about training errors also hold for validation errors. So, to restate, copycat problems occur when the changepoint fraction is lower than the error of the reward-optimal imitator.

While the above argument is not fully rigorous or comprehensive,\footnote{In particular, Eq~\eqref{eq:copycat_condition} does not rule out that that may be other parameter vectors $\theta \neq \theta^{R*}$ that yield higher reward and produce lower error than the copycat $\psi^*(\cdot)$.} it yields strong intuitions for the factors that make copycat problems more likely in POMDP imitation: \textbf{(1)} the higher the temporal correlation among expert actions, the more infrequent the changepoints (i.e., lower $\epsilon_{CP}$), and \textbf{(2)} the harder the imitation setup (such as high-dimensional observations or noisy demonstrations), the higher the value of $\text{MSE}_\mathcal{D}(\theta^{R*})$. In both cases, the copycat-producing inequality in Eq~\eqref{eq:copycat_condition} becomes more likely to hold.

These observations directly motivate our approach. 
We assume fixed standard datasets, architectures, and optimizers in this paper, so we cannot address \textbf{(2)} above.
However, we can artificially inflate the changepoint fraction $\epsilon_{CP}$ simply by upweighting changepoints when setting up the BC objective, to address \textbf{(1)}. This is the crux of our approach, which we describe in more detail next.

\subsection{Reweighted Behavioral Cloning Objective}\label{sec: reweight-by-mutation}

In datasets with high temporal correlation among expert samples, the natural changepoint fraction $\epsilon_{CP}$ is very low, which makes copycat issues more likely, as we have argued above. However, we can effectively upsample these changepoints by shifting to a \emph{weighted} version of the behavioral cloning objective in Eq~\eqref{eq:bc_loss}:
\begin{equation}\label{eq: w-weight-obj}
    \theta^{*} = \mathop{\arg\min}_{\theta}\sum_{t=0}^N w_{t}(\pi_{\theta}(\tilde{o}_t) - a_{t})^{2},
\end{equation}
where $w_{t}$ is the weight for each data sample. With the right weighting scheme, the reweighted MSE error of the copycat policy $\epsilon_{CP}$ would rise and therefore making the condition in Eq~\eqref{eq:copycat_condition} more difficult to meet, alleviating the copycat problem.

\subsection{Action Prediction Error (APE)}\label{sec: reweight-by-APE}

\begin{figure}[t]
    \centering
    \includegraphics[width=0.47\textwidth]{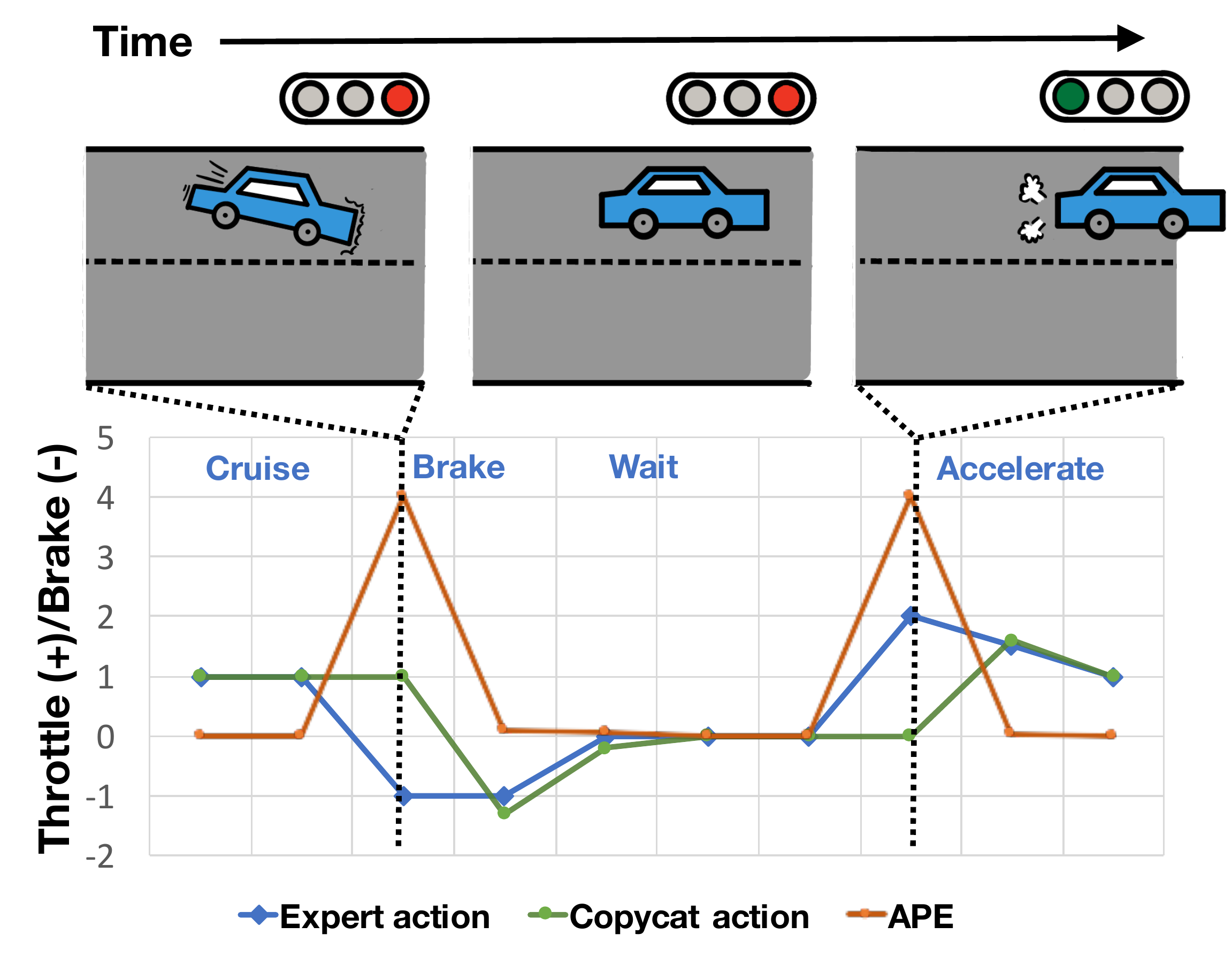}
    \caption{Action prediction error (APE) computation (Equation~\eqref{eq:APE}) as a function of expert and copycat actions in a traffic light setting similar to Figure~\ref{fig: copycat}. APE peaks align with keyframes.}.
    \label{fig: ape}
\end{figure}

What would an appropriate sample weighting scheme look like? Since the copycat problem arises from exploiting temporal correlation among expert actions, we must upweight and emphasize those keyframe samples where this correlation breaks down. Identifying such samples amounts to a type of changepoint detection in the expert action sequence. While many generic changepoint and keyframe detection approaches have been proposed for time series or video~\cite{burg2020evaluation,sheng2019unsupervised}, in our specialized setting, the most appropriate choice is a changepoint detection score that is closely tied to the copycat policy defined in Eq~\eqref{eq:copycat_loss}, as foreshadowed in Sec~\ref{sec: motivation}. Specifically, we first train a small MLP copycat policy network $\psi^*$ with the training objective of Eq~\eqref{eq:copycat_loss} --- recall that the only inputs to this policy are the past actions $[a_{t-1}, a_{t-2}, \cdots ]$. Then, we use its prediction error for each training sample to set the sample weight $w_t$ in the reweighted BC objective of Eq~\eqref{eq: w-weight-obj}. These weights need only be computed once, before training the BC policy $\pi_\theta(\tilde{o}_t)$.

In more detail, for every training sample $(\tilde{o}_t, a_t)\sim\mathcal{D}$, we define the ``action prediction error" (APE) as the squared error of the copycat policy $\psi^*$ with respect to expert actions:
\begin{equation}
    \text{APE}_t = (\psi^* (a_{t-1}, a_{t-2}, \cdots) - a_{t})^{2}. \label{eq:APE}
\end{equation}
Figure~\ref{fig: ape} shows a schematic. To avoid copycat overfitting when working with small datasets, the APE can instead be computed through cross-validation, always training copycat policies and measuring their errors on disjoint data. 

\begin{figure}[t]
    \centering
    \includegraphics[width=0.45\textwidth]{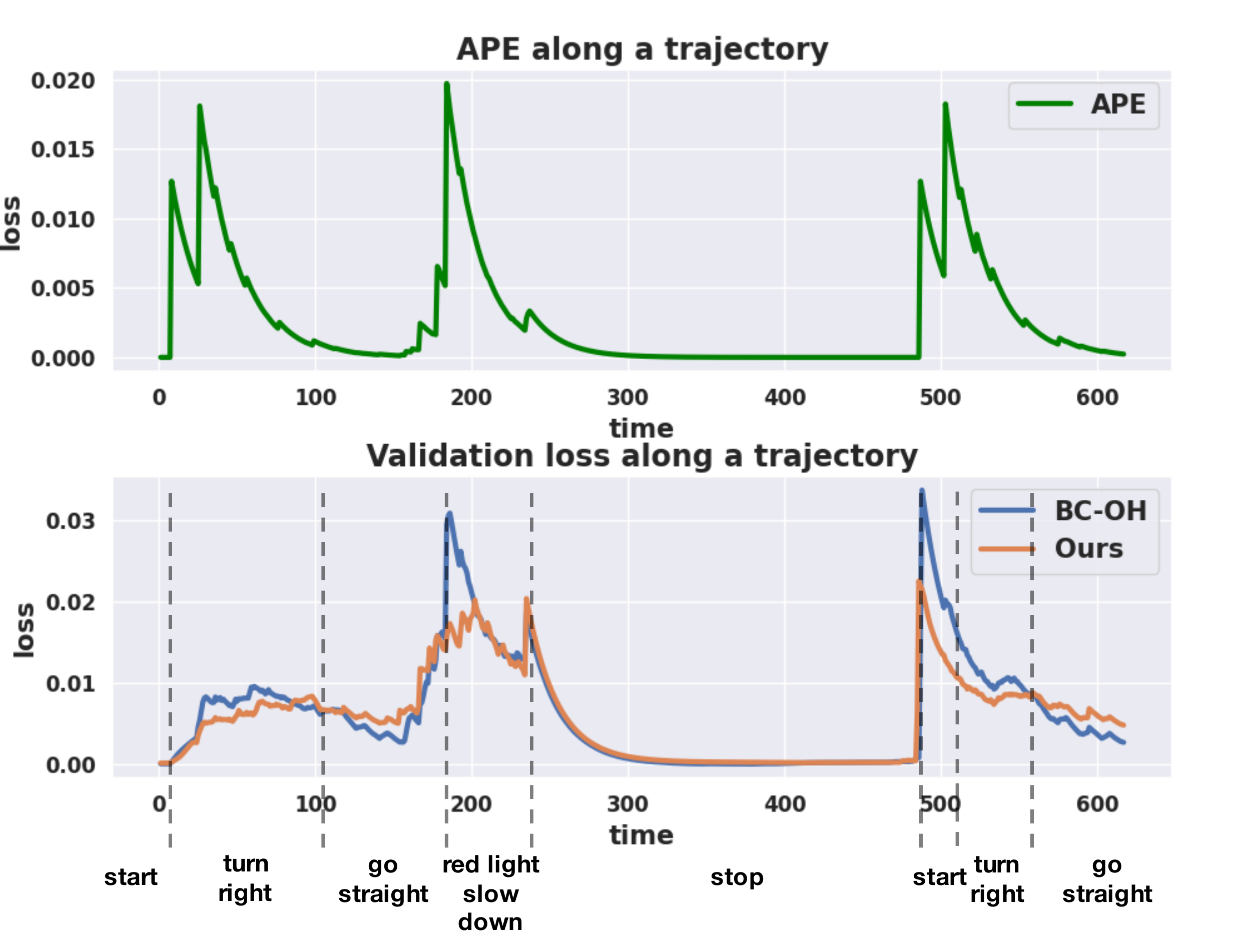}
    \caption{Action prediction error (APE) and behavioral cloning loss of BC-OH and our approach along a validation driving trajectory in CARLA. Dotted lines segment the trajectory into different annotated phases. APE is well-aligned with BC-OH errors.
    }
    \label{fig: carla_loss_along_traj}
\end{figure}

Samples with high $\text{APE}_t$ are more likely to be changepoints. Since we would like to upweight changepoints, we set the sample weights $w_t$ in Eq~\eqref{eq: w-weight-obj} to be monotic non-decreasing functions of $\text{APE}_t$: 
\begin{equation}
w_t = f(\text{APE}_t). \label{eq:weight-setting}
\end{equation}
Figure~\ref{fig: carla_loss_along_traj} shows a plot of the APE and the MSE loss for BC-OH along a validation trajectory, for a driving policy in a photorealistic image-based driving environment. Samplewise BC-OH errors align very well with the APE, which are the copycat errors, verifying the existence of the copycat issue. We evaluate setting $f(.)$ to $\rm{softmax}$ and $\rm{step}$ functions in our experiments. Plugging this back into Eq~\eqref{eq: w-weight-obj}, all that remains is to train the BC policy by solving:
\begin{equation}
    \theta^{*} = \mathop{\arg\min}_{\theta}\sum_{t=0}^N f(\text{APE}_t)(\pi_{\theta}(\tilde{o}_t) - a_{t})^{2}.
\end{equation}
Algorithm~\ref{alg: ours} summarizes our complete approach. Intuitively, our approach amounts to focusing the behavioral cloning objective on the demonstration frames where copycat policies fail, so that the BC learner becomes less likely to discover such copycat policies.

\begin{algorithm}[tb]
  \caption{Keyframe-Focused Visual Imitation Learning}
  \label{alg: ours}
\begin{algorithmic}[1]
  \STATE {\bfseries Input:} Expert demonstrations $\mathcal{D}=\{(\tilde{o}_{t}, a_{t)}\}$.

  \STATE Train an optimal copycat policy MLP $\psi^{*}$ on $\mathcal{D}$ (Eq~\eqref{eq:copycat_loss}).\\
  \STATE Compute APE$_t$ for each sample in $\mathcal{D}$ (Eq~\eqref{eq:APE}).\\
  \STATE Compute the sample weights $w_t$ (Eq~\eqref{eq:weight-setting}).\\
  \STATE Optimize the imitation policy neural net $\pi_{\theta}$ to minimize the reweighted behavioral cloning objective (Eq\eqref{eq: w-weight-obj}).\\
  \STATE {\bfseries Return} $\pi_{\theta}$
\end{algorithmic}
\end{algorithm}

\subsection{Implementation Details}
Our copycat policy network $\psi^{*}$ is a a two-layer MLP.
For $f(\cdot)$, we experiment with $\rm{softmax}$ and $\rm{step}$ functions. The $\rm{softmax}$ function is applied within each training minibatch, i.e.,  $w_{i}=\frac{e^{\tau\text{APE}_{i}}}{\sum_{j}e^{\tau\rm{APE}_{j}}}$; the temperature $\tau$ is a hyper-parameter. The $\rm{step}$ function assigns a constant weight $w_{i}=W$ to samples in the top $\rm{THR}$ percentile of APE and $w_i=1$ otherwise; $W$ and $\rm{THR}$ are hyper-parameters. All hyperperameters were set through a simple grid search; see Supp for details.
All policies using observation histories are trained by stacking image observations along the channels dimension. Architectural details are environment-specific and discussed in Sec~\ref{sec: experimental_setup}.

Our method introduces barely any computational overheads over baseline behavioral cloning (BC-OH). At test time, our method is exactly identical to BC-OH. At training time, the only extra steps are training the copycat policy $\psi^{*}$ and calculating the sample weights, before following the BC-OH training procedure. Since the inputs to $\psi^*$ are only the action histories, rather than the visual observations, this all amounts to a fast data preprocessing step (less than 15 mins even on our largest and most complex environments). Further, once this is completed, any number of policies may be trained on that data with zero additional overhead.

\section{Experimental Setup}\label{sec: experimental_setup}

We now comprehensively evaluate our approach on a photorealistic driving simulator, CARLA~\cite{Dosovitskiy17}, and three image-based OpenAI Gym MuJoCo robotics environments.

\begin{figure}[t]
    \centering
    \includegraphics[width=0.136\textwidth]{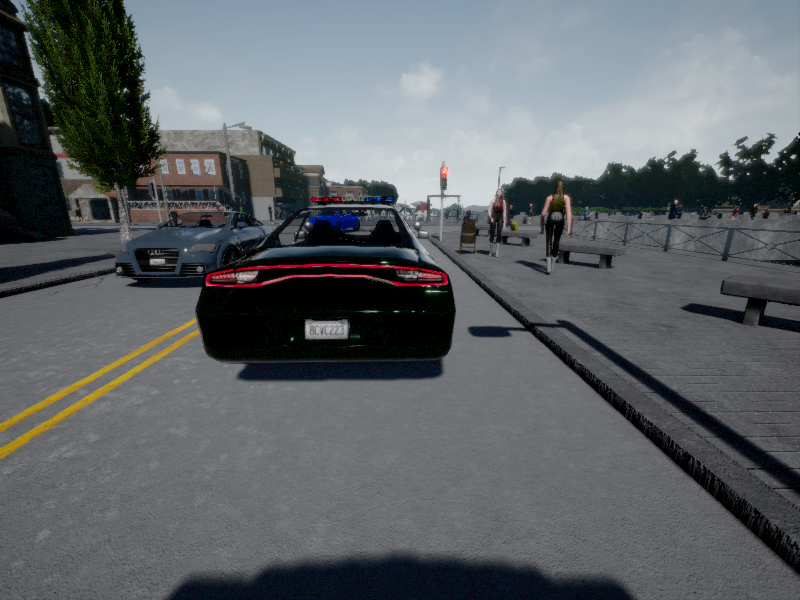}
    \includegraphics[width=0.102\textwidth]{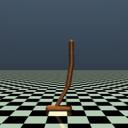}
    \includegraphics[width=0.102\textwidth]{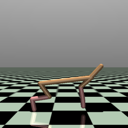}
    \includegraphics[width=0.102\textwidth]{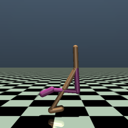}
    \caption{The four environments used in our experiments: CARLA, Hopper, HalfCheetah and Walker2d.}
    \label{fig: env-demo}
\end{figure}

\textbf{CARLA.} CARLA is a photorealistic urban driving simulator with varying road and traffic conditions. It has recently emerged as a standard testbed for visual imitation learning, through the publicly available 100-hour CARLA100 driving dataset~\cite{codevilla2019exploring}. This dataset is generated by a PID expert controller with access to simulator states.
We use the hardest CARLA100 benchmark, \textit{NoCrash-Dense}, which has the most pedestrians and traffic. We set history size $H=6$. For each method, we train three policies from random initializations, and evaluate each policy three times to account for environmental stochasticity. We measure the mean and standard deviation of four metrics: \textit{\%success}, \textit{\#collision}, \textit{\%progress} and \textit{avg.~speed}. \%success is the number of test episodes correctly completed by the agent, \#collision counts the times the agent crashes into pedestrians, vehicles and other obstructions, \%progress measures the fraction of the distance traveled towards a goal location, and avg.~speed is the average speed at which the agent drives. All metrics are measured on 100 predefined benchmark test episodes. More details in Supp.

Note that CARLA100 is a particularly challenging testbed because it applies the best known techniques for alleviating distributional shift issues in offline imitation, namely, noise injection~\citep{laskey2017dart} which is an offline counterpart of DAGGER~\cite{Ross2011}, and multi-camera data augmentation~\cite{bojarski2016end,giusti2015machine}. Further, all approaches use the speed prediction regularization scheme introduced in ~\citet{codevilla2019exploring} to partially address the copycat problem (coined there as the ``inertia problem''). Finally, we train all approaches with Imagenet-pretrained Resnet-34 backbones~\cite{codevilla2019exploring} and weighted control losses~\citep{codevilla2018end} to reflect the state of the art. See Supp. More broadly, autonomous driving is the setting in which prior works have most often reported severe copycat issues~\cite{muller2006off,bansal2018chauffeurnet,codevilla2019exploring,wang2019monocular}. Any persistent copycat issues in CARLA thus represent a key open problem in imitation learning.

\textbf{MuJoCo-Image (Hopper, HalfCheetah, Walker2D).} Following previous work that had identified environments where the copycat problem arises~\cite{NIPS2019_9343,chuan2020fighting}, we evaluate our method in three standard OpenAI Gym MuJoCo continuous control environments: Hopper, HalfCheetah and Walker2D. We set the observation $o_t$ to be the 128x128 RGB image of the environment, naturally excluding velocity and force information and making the environments partial observed. We set the history size $H=1$, so that $\tilde{o}_t=[o_{t-1}, o_t]$. These tasks vary in their state and action spaces, environmental dynamics, and reward structure. We generate expert data from a TRPO policy~\cite{schulman2015trust} 
with access to true states (1k samples for HalfCheetah, and 20k for Hopper and Walker2D).
For each imitation method, we train three policies from random initializations and report the reward mean and standard deviation. 
See Supp for hyperparameters and training details.

\subsection{Baselines and Ablations}

We compare our method against the following baselines:

\textbf{Behavioral Cloning (BC-SO and BC-OH).}
As introduced in Sec~\ref{sec: introducing-copycat}, BC-SO and BC-OH are BC with a single observation and observation histories respectively.

\textbf{HistoryDropout.} \citet{bansal2018chauffeurnet} proposed to randomly drop out the historical part of the observations to tackle copycat problems in imitation for driving.
We implement this baseline by adding a dropout layer to the past observations, i.e. $o_{t-1}, o_{t-2}, \cdots$.

\textbf{Fighting-Copycat-Agents (FCA).}
\citet{chuan2020fighting}
proposed to remove all information about the last action $a_{t-1}$ from an embedding of the observation history, using adversarial learning. They report promising results in low-dimensional state-based environments, and we extend their publicly available code to our image-based settings, with upgraded backbone networks and re-tuned hyperparameters. See Supp for details.

\textbf{DAGGER.} This is a widely used method to mitigate distributional shift issues in imitation learning~\cite{Ross2011}. While our method operates completely offline, DAGGER requires online environmental interaction with a queryable expert. Nevertheless, it provides a useful comparison point. We set the number of environment queries to 100 and 1k for the MuJoCo environments and 120K for CARLA.

We also attempted to compare against \citet{NIPS2019_9343}, which, like DAGGER, proposes an online approach that targets ``causal confusion'', a more general version of the copycat problem. However, their causal graph learning method, demonstrated with up to 30 observation dimensions at most, does not scale to our image-based settings.

Aside from these standard and published baselines for imitation learning, we also study three ablations of our approach, replacing our APE-based sample reweighting with alternatives: (1) \textbf{BCPD}~\cite{xuan2007modeling} represents the widely used family of Bayesian changepoint detection techniques~\cite{adams2007bayesian,fearnhead2006exact} for general multivariate time series, (2) \textbf{ActFreq} clusters expert actions in the training data to form action ``categories'' before applying category frequency-based sample reweighting, a standard approach for handling imbalanced data~\cite{DBLP:journals/corr/abs-1106-1813, DBLP:journals/corr/abs-1712-03162,cui2019class,cao2019learning,kang2019decoupling,zhou2020bbn}, and (3) \textbf{Boosting} uses the standard Adaboost~\cite{freund1997decision} scheme for iteratively training BC-OH policies and upweighting high error samples. See Supp for more details about these ablations.

\section{Results and Analysis}~\label{sec:results}
We now report the results of experiments performed to answer: \textbf{(1)} Does our method improve visual behavior cloning from observation histories? \textbf{(2)} Does it handle changepoints well? \textbf{(3)} To what extent does it reduce distributional shift issues in the learned policies? \textbf{(4)} Do our policies behave less like copycat policies?, and \textbf{(5)} When do our policies perform worse than the baselines?

\begin{question}
Does our method improve visual behavior cloning from observation histories?
\end{question}

\textbf{\textit{CARLA}.~~~} See Table~\ref{tab: carla-withspeed} for \%success results, and Supp for other metrics.
The single-frame imitator BC-SO performs significantly better than BC-OH, illustrating the copycat problem. Our method easily outperforms all history-based baselines, including, surprisingly, even DAGGER which has the advantage of 120k expert queries! As we show in Supp, DAGGER does drive at higher average speed (18.5 km/h vs.~14.9 km/h), but at the cost of many more collisions (60 vs.~43) than our method. On other metrics (\#collision and \%progress), our method is comfortably best. Of the three sample reweighting ablations, BCPD performs the best, but still produces worse results compared to BC-OH without any sample reweighting, and falls far short of our approach.

\begin{figure*}[t]
    \centering
    \includegraphics[width=0.95\textwidth]{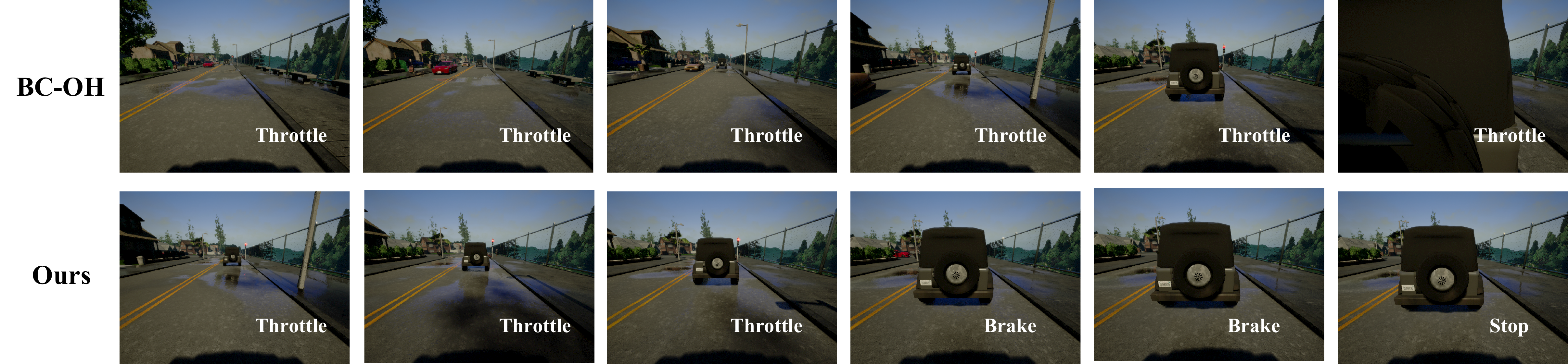}
    \caption{
    An example test scenario from CARLA navigated by a BC-OH policy (top) and ours (bottom). Frames displayed in sequence from left to right. Actions (``Throttle"/``Brake") are overlaid on the frames. 
    }
    \label{fig: carla-qualitative}
\end{figure*}

However, even with these large gains over history-based baselines, our approach only recovers the performance of the single-frame imitator BC-SO --- we do not significantly surpass it. Specifically, we get comparable \%success, \%progress, and avg.~speed with higher consistency (lower variance) and fewer \#collisions. We believe the limited extent of these gains may be because this setting does not emphasize information integration over time. CARLA-100 data~\cite{codevilla2019exploring} is collected largely in low traffic settings where the ego-agent’s own speed might very well be the main historical information missing in the current image observation. However, CARLA-100 provides the velocity as part of the observation, i.e., $o_t=[\text{image}_t, \text{velocity}_t]$. Thus, BC-SO already has access to agent velocity, meaning that the environment is nearly fully observed.

\textbf{\textit{CARLA-w/o-speed}.~~~} To understand why the CARLA setting does not reward agents that condition on multiple frames,
we report results in a modified setting, CARLA-w/o-speed, where we withhold the ego-agent velocity from the observation for all methods. See Table~\ref{tab: carla-withspeed} (right). The main differences from above are: (1) BC-SO is dramatically worse than before, (2) BC-OH improves significantly over BC-SO, and (3) our method improves by a large margin over BC-OH and therefore over BC-SO. These findings suggest that the ego-agent velocity does indeed encapsulate most of the driving-relevant information contained in $\tilde{o}_t$, and our method makes significant progress towards recovering this information from the frame history. Our approach continues to comprehensively outperform all history-based baselines. All three alternative sample reweighting schemes all continue to perform poorly in this setting. See Supp for other metrics, which are consistent with the above results.

Figure~\ref{fig: carla-qualitative} shows an example test sequence from CARLA where BC-OH speeds straight into a slow-moving car in front of it, while our policy correctly slows down as the car nears, to avoid crashing. 

\begin{table}[t]
\caption{CARLA \%success ($\uparrow$). More metrics in Supp.}
\label{tab: carla-withspeed}
\vskip 0.15in
\begin{center}
\begin{small}
\begin{sc}
\resizebox{0.48\textwidth}{!}{  
\begin{tabular}{ccc}
\toprule
Method & CARLA & CARLA-w/o-speed \\
\midrule
BC-SO    &  \textbf{42.667 $\pm$ 8.668} & 9.222 $\pm$ 2.380 \\
BC-OH &  33.000 $\pm$ 4.190   & 25.667 $\pm$ 0.981\\
Ours (step)    &  \textbf{43.444 $\pm$ 0.786}  & \textbf{36.778 $\pm$ 5.808} \\
\midrule
FCA     &  35.667 $\pm$ 3.559 &  27.444 $\pm$ 4.113\\
HistoryDropout  &  34.000 $\pm$ 2.625 &  25.333 $\pm$ 5.375 \\
DAGGER (120K)   &  35.222 $\pm$ 3.067  & 28.333 $\pm$ 3.496 \\
\midrule
BCPD   &  28.667 $\pm$ 2.494  & 20.000 $\pm$ 1.414 \\
ActFreq   &  20.333 $\pm$ 5.825  & 14.667 $\pm$ 1.764 \\
Boosting   &  3.000 $\pm$ 1.414  & 10.0 $\pm$ 2.160 \\
\bottomrule
\end{tabular}
}
\end{sc}
\end{small}
\end{center}
\vskip -0.1in
\end{table}

\textbf{\textit{Hopper, HalfCheetah, Walker2D}.~~~} See Tab~\ref{tab: mujoco-nospeed}. BC-OH does manage to yield higher rewards than BC-SO in these settings, but it is further improved by addressing the copycat problem.
We experimented with two simple choices of monotonic transformations $f(\cdot)$ applied to APE in Eq~\eqref{eq:weight-setting}, namely $\rm{step}$ and $\rm{softmax}$ --- $\rm{softmax}$ performs consistently better. While $\rm{step}$ is arguably a simpler weighting scheme, we find that $\rm{softmax}$ enjoys the benefits of easier hyperparameter tuning since it requires only a single temperature hyperparameter. Compared to all the other offline baselines, Ours (Softmax) easily yields the highest rewards across all environments. With the advantage of online interaction and expert queries, DAGGER with 100 queries is worse than our method on Hopper and Walker2D but better on HalfCheetah. With 1k queries, DAGGER performs marginally better than our method on all three environments.

\begin{table}[t]
\caption{MuJoCo-Image environment rewards ($\uparrow$).}
\label{tab: mujoco-nospeed}
\vskip 0.15in
\begin{center}
\begin{small}
\begin{sc}
\resizebox{0.48\textwidth}{!}{
\begin{tabular}{ccccc}
\toprule
Method & Hopper & HalfCheetah &  Walker2D \\
\midrule
BC-SO            & 601$\pm$168   &       4 $\pm$ 5   &   481$\pm$40     \\
BC-OH            & 740$\pm$35    &       615 $\pm$ 41  &   614 $\pm$ 107     \\
Ours (step)      & 905$\pm$135            &  470 $\pm$ 205  &  654 $\pm$53     \\
Ours (Softmax)   & \textbf{951$\pm$117} &   \textbf{819$\pm$96} &  \textbf{769$\pm$97}      \\
\midrule
FCA              &     735$\pm$ 106          &   270 $\pm$ 168  &   534 $\pm$ 99   \\
HistoryDropout       & 617$\pm$111   &     96$\pm$40          & 594$\pm$61\\
DAGGER (100)       & 745$\pm$157   &  936 $\pm$ 86  & 598$\pm$26       \\
DAGGER (1k)      & 1034$\pm$45   &  822 $\pm$ 186  &   699 $\pm$ 111     \\
\bottomrule
\end{tabular}
}
\end{sc}
\end{small}
\end{center}
\vskip -0.1in
\end{table}

\begin{question}
Does our method imitate the expert better at the changepoints, as it was designed to do? What about non-changepoints?
\end{question}
We showed earlier in Figure~\ref{fig: carla_loss_along_traj} that high APE samples (i.e., changepoints) do in fact correspond well with imitation errors in BC-OH models. The same figure also plots the error for our approach. At all the changepoints, such as turning right and slowing down in front of a red light, the validation errors of our method are significantly lower than BC-OH. On other frames, it sometimes produces higher errors than BC-OH.

\begin{figure}
    \centering
    \includegraphics[width=0.5\textwidth]{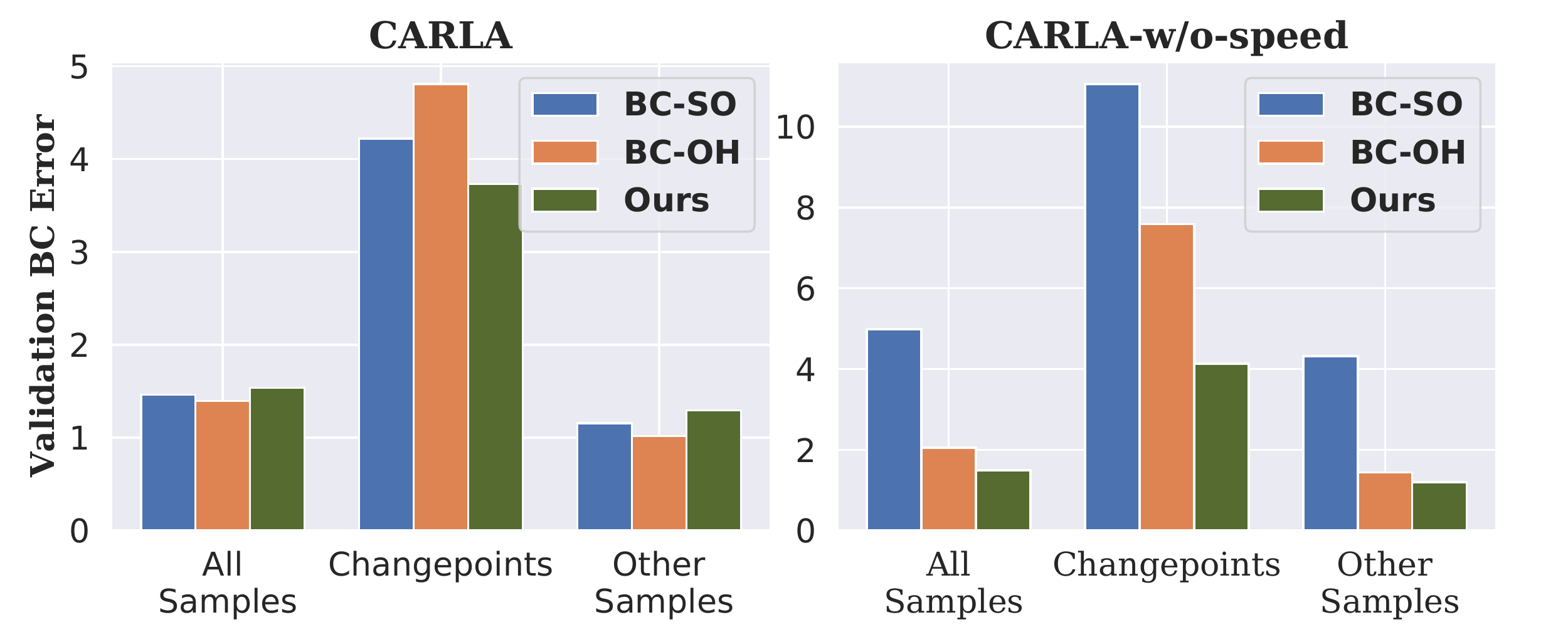}
    \caption{Imitation MSE losses for different sets of validation frames: changepoints, others, and all combined.}
    \label{fig: loss-info-chart}
\end{figure}

We investigate this phenomenon more quantitatively, reporting the unweighted imitation MSE (corresponding to Eq~\eqref{eq:bc_loss}) for BC-SO, BC-OH, and ours, on all frames, and then separately for changepoints and other frames. All MSEs are computed on held-out data. See Figure~\ref{fig: loss-info-chart}. On CARLA, BC-OH performs worse than BC-SO at the APE-based changepoints and better at other frames, once again validating our changepoint-focused approach. In comparison, our approach performs significantly better on changepoints and marginally worse on other samples. Since there are many fewer changepoints, this corresponds to marginally higher overall validation MSE for our approach (despite the higher reward). This is not directly a concern however since, as we have reported above, our method comprehensively outperforms BC-OH in terms of environment reward. Instead, this finding lines up with our intuition that, for optimal reward, it is more important to act correctly at some ``keyframes'' than at others. Supp has full quantitative results.

Finally, on CARLA-w/o-speed, BC-SO suffers from removing agent velocity information, producing the highest errors on changepoints as well as other samples. Our method yields the lowest errors in both cases (and therefore lowest overall). Note that while we report validation losses here, Supp shows very similar trends for training losses. We find this trend surprising: BC-OH is trained on the unweighted loss over all samples and yet produces a policy that has higher value of this loss on the training set than our approach which optimizes a weighted objective that emphasizes changepoints. We believe this may be a case of data rebalancing avoiding \emph{optimization}-related shortcuts. Similar phenomena have been observed before in ML systems that amplify biases in the training data~\cite{geirhos2020shortcut}.

\begin{question}
To what extent does our method reduce distributional shift issues in the imitation policies?
\end{question}

For each policy, we now compute the BC MSE of Eq~\eqref{eq:bc_loss} on the \emph{test data} generated by executing the policy in the environment. The resulting ``rollout imitation error'' directly measures distributional shift between the expert data and the policy. Note that this measurement is possible in CARLA because the CARLA expert is rule-based rather than learned from data --- therefore, it does not itself suffer from distributional shift, and allows evaluating shift issues with respect to the imitator alone. On both CARLA and CARLA-w/o-speed, our policies (0.07, 0.16) have lower rollout imitation error than BC-OH (0.11, 0.40). This suggests that our approach does in fact suffer less distributional shift.

\begin{question}
Do our policies behave less like copycats?
\end{question}

We have thus far measured APE by training copycat policies on expert action sequences and measuring their errors. We now define a similar notion called $\text{APE}(\pi)$. To measure $\text{APE}(\pi)$ for some policy $\pi$, we generate data $\mathcal{D}_\pi$ by executing $\pi$, then train a new optimal copycat policy $\psi^*_\pi$ on $\mathcal{D}_\pi$, and measure its average error on held-out data (generated from $\pi$ again).
This ``avgAPE$(\pi)$" measures how temporally correlated actions from $\pi$ tend to be --- lower error corresponds to less interesting policies that generate smooth, predictable action sequences. \citet{chuan2020fighting} used a similar metric and showed that approaches that suffer from the copycat problem commonly have lower avgAPE$(\pi)$ than the expert policy. This is related to our comment above about bias amplification and shortcuts: if the expert policy has low avgAPE$(\pi)$, the imitator trained to mimic it ends up with even lower avgAPE$(\pi)$.

Our results, shown in Table~\ref{tab: roll-out-APE}, are consistent with this. BC-OH has much lower avgAPE$(\pi)$ than the expert in all environments. Our method consistently improves upon BC-OH, but continues to produce lower avgAPE$(\pi)$ than the expert. While this is not a direct metric, it indicates that our method makes progress towards resolving copycat policy learning, and that there may still be further room for improvement.

\begin{table}[t]
\caption{avgAPE for various approaches. All values are ($\times 10^{-2}$)}
\label{tab: roll-out-APE}
\vskip 0.15in
\begin{center}
\begin{small}
\begin{sc}
\resizebox{0.48\textwidth}{!}{
\begin{tabular}{ccccccc}
\toprule
Method & Carla & CarlaNS &  Hopper & HalfCheetah &  Walker2D \\
\midrule
Expert &  1.602  &  1.602  &  0.86  &  9.81  & 2.47   \\\midrule
BC-OH &  0.966  &  0.741  & 0.61   & 5.86 &    0.74   \\
Ours    &  \textbf{1.187}  &  \textbf{1.305}  &  \textbf{0.75}  &   \textbf{9.00}    &  \textbf{0.85} \\
\bottomrule
\end{tabular}
}
\end{sc}
\end{small}

\end{center}
\vskip -0.1in
\end{table}

\begin{question}
When does keyframe-focused imitation perform systematically worse than simple behavior cloning baselines?
\end{question}

Our approach specifies that weights for frames in training data should be set as monotonic functions $f$ of the action prediction error of a copycat policy, as specified in Eq~\ref{eq:weight-setting}. In practice, the choice of the weighting function $f$ is important. While experimenting with various options for $f$, we observed that overly flat functions $f$ would not sufficiently alter the behavior of BC-OH, but overly steep functions would sometimes assign inordinately large weights to changepoint keyframes, causing the model to underfit to ordinary frames which constitute the majority of the data. For example, when $f$ is set to the step function, with a high value $W$ assigned to high-APE frames, that trained policy fails even to follow its lane sometimes. 
In our experimental setups, we found that the sweet spot of functions $f$ that produced our desired behavior was easy to find through a search over the parameters of simple function families ($\rm{step}$ and $\rm{softmax}$). Further, the same parameters worked well across many setups. We report hyperparameter sensitivity results in Supp.

Another potential failure case is in imitation datasets where some samples have noisy action labels. Upweighting changepoints using our approach might assign high weights to such samples, since the copycat policy would fit the noise poorly. Eventually, this might produce bad policies. While we haven't encountered this in our experiments, it might warrant systematic study in future work.

\section{Conclusion}
We have proposed a sample weighting strategy to learn effective imitation policies that can integrate information over time without succumbing to learning copycat shortcut policies, while also being very easy to implement and tune. Stepping back to take a broader view, our results show that minimizing the standard empirical risk as in Eq~\eqref{eq:bc_loss} is not optimal in offline imitation learning because of distributional shift issues. Instead, minimizing a carefully reweighted empirical risk produces better-performing policies. 

Across four image-based environments spanning simulated locomoting robots and photorealistic urban driving, our approach trivially scales well and yields better results than all prior approaches tackling similar issues. On the long-standing problem of behavioral cloning for driving, we demonstrate that the widely used current standard benchmark CARLA100 might not be challenging enough to effectively benefit from information integration across time, and show large gains in a modified variant that does require such information integration. A future benchmark with more unpredictable vehicles, pedestrians, congested roads, and obstructions would offer a more realistic evaluation of current approaches.

\section{Acknowledgement}
This work is supported by an Amazon Research Award and gift funding from NEC Laboratories America to DJ, and funding from the Ministry of Science and Technology of the People's Republic of China, the 2030 Innovation Megaprojects "Program on New Generation Artificial Intelligence" (Grant No. 2021AAA0150000) to YG.

\bibliography{main}
\bibliographystyle{plainnat}

\newpage

\section{Appendix}

\begin{table*}[t]
\caption{CARLA test results in Nocrash-Dense benchmark. BC-SO is significantly better than BC-OH, verifying the causal confusion phenomenon. Our method outperforms both BC-SO and BC-OH with higher success rate. Moreover, our method significantly reduces the \#collision, which is very important for urban driving tasks. Comparing with the baselines, we do better in all the four metrics than the two offline algorithms and even beats the Dagger which require online query.}

\label{tab: carla-withspeed}
\vskip 0.15in
\begin{center}
\begin{small}
\begin{sc}
\begin{tabular}{ccccc}
\toprule
Method & \%success (/100) ($\uparrow$) & \#collision ($\downarrow$)  & \%progress ($\uparrow$)  &  avg.~speed ($\uparrow$)  \\
\midrule
BC-SO~\citep{codevilla2019exploring}    &  \textbf{42.667 $\pm$ 8.668}  &  48.444 $\pm$ 9.044  &  \textbf{0.580 $\pm$ 0.055}  &  15.559 $\pm$ 3.035  \\
BC-OH &  33.000 $\pm$ 4.190  &  52.111 $\pm$ 5.878  &  0.497 $\pm$ 0.042  &  11.775 $\pm$ 3.225  \\
Ours (step)    &  \textbf{43.444 $\pm$ 0.786}  &  \textbf{42.615 $\pm$ 2.228} &  \textbf{0.580 $\pm$ 0.040}  &  14.938  $\pm$ 2.759  \\
\midrule
FCA~\citep{chuan2020fighting}     &  35.667 $\pm$ 3.559  &  50.333 $\pm$ 4.643  &  0.551 $\pm$ 0.030  &  13.927 $\pm$ 2.905  \\
HistoryDropout~\cite{bansal2018chauffeurnet}  &  34.000 $\pm$ 2.625  &  60.222 $\pm$ 3.119  &  0.506 $\pm$ 0.029  &  17.847 $\pm$ 2.120  \\
DAGGER 120K~\cite{Ross2011}   &  35.222 $\pm$ 3.067  &  60.000 $\pm$ 2.625  &  0.512 $\pm$ 0.026  &  \textbf{18.515 $\pm$ 1.586}  \\
BCPD &  28.667 $\pm$ 2.494  &  36.667 $\pm$ 8.260  &  0.440 $\pm$ 0.057  &  10.071 $\pm$ 6.668 \\
ActFreq &  20.333 $\pm$ 5.825  &  26.000 $\pm$ 5.354  &  0.410 $\pm$ 0.089  &  9.229 $\pm$ 3.353  \\
Boosting &  3.000 $\pm$ 1.414  &  19.333 $\pm$ 4.110  &  0.101 $\pm$ 0.026  &  1.933 $\pm$ 2.174 \\
\bottomrule
\end{tabular}
\end{sc}
\end{small}
\end{center}
\vskip -0.1in
\end{table*}

\begin{table*}[t]
\caption{CARLA-w/o-speed results in Nocrash-Dense benchmark. In this case, BC-SO's performance is much worse than BC-OH for lack of historical information and it is pretty  difficult to infer speed from a single frame. Our method still performs well in such an information-constrained situation.}
\label{tab: carla-nospeed}
\vskip 0.15in
\begin{center}
\begin{small}
\begin{sc}
\begin{tabular}{ccccc}
\toprule
Method & \%success (/100) ($\uparrow$) & \#collision ($\downarrow$)  & \%progress ($\uparrow$)  &  avg.~speed ($\uparrow$)  \\
\midrule
BC-SO~\cite{codevilla2019exploring}    &  9.222 $\pm$ 2.380  &  84.667 $\pm$ 4.769  &  0.204 $\pm$ 0.033  &  35.182 $\pm$ 1.594  \\
BC-OH &  25.667 $\pm$ 0.981  &  60.889 $\pm$ 1.911  &  0.467 $\pm$ 0.049  &  15.543 $\pm$ 3.714  \\
Ours (step)    &  \textbf{36.778 $\pm$ 5.808}  &  \textbf{45.333 $\pm$ 4.690}  &  \textbf{0.540 $\pm$ 0.050}  &  14.416 $\pm$ 3.145  \\
\midrule
FCA~\citep{chuan2020fighting}     &  27.444 $\pm$ 4.113  &  59.222 $\pm$ 5.996  &  0.453 $\pm$ 0.052  &  13.856 $\pm$ 2.729  \\
HistoryDropout~\citep{bansal2018chauffeurnet}    &  25.333 $\pm$ 5.375  &  66.778 $\pm$ 5.865  &  0.449 $\pm$ 0.047  &  16.086 $\pm$ 3.889  \\
DAGGER 120K~\citep{Ross2011}  &  28.333 $\pm$ 3.496  &  62.667 $\pm$ 3.771  &  0.457 $\pm$ 0.014  &  \textbf{16.988 $\pm$ 2.757}      \\
BCPD &  20.000 $\pm$ 1.414  &  44.000 $\pm$ 9.899  &  0.400 $\pm$ 0.0626  &  5.379 $\pm$ 2.259  \\
ActFreq &  14.667 $\pm$ 1.764  &  24.667 $\pm$ 3.972  &  0.313 $\pm$ 0.031  &  3.649 $\pm$ 3.073  \\
Boosting &  10.000 $\pm$ 2.160  &  49.667 $\pm$ 4.028  &  0.223 $\pm$ 0.026  &  12.502 $\pm$ 3.242  \\
\bottomrule
\end{tabular}
\end{sc}
\end{small}
\end{center}
\vskip -0.1in
\end{table*}

\subsection{Additional Details on CARLA Experiments}
In Sec 5, we specified the experimental setup for our experiments in the CARLA driving environment, which closely follows standard protocol from~\citet{codevilla2018end,codevilla2019exploring,chen2020learning}. We provide additional details here.

\paragraph{Data Collection.} The CARLA100 dataset~\citep{codevilla2019exploring} in our experiments is collected with noise injection~\cite{laskey2017dart} to perturb 10\% of expert actions.
It also uses three cameras: a forward-facing one and two lateral cameras facing 30 degrees away towards left or right~\cite{bojarski2016end}, for data augmentation to guard against distributional shift. 

\paragraph{Architectures and Training Details.} All our baseline models use ImageNet-pretrained Resnet34 as our perception model~\cite{codevilla2019exploring}. We use the state-of-the-art conditional imitation learning model \textit{CILRS}~\citep{codevilla2019exploring} as our backbone, with a weighted control loss~\cite{codevilla2018end} that assigns weights of 0.5 to steer, 0.45 to throttle and 0.05 to brake. We train all models for $10^{5}$ training iterations with minibatch size 120. We use Adam optimizer, set the initial learning rate to $1 \times 10^{-4}$ and decay the learning rate by $0.1$ whenever the loss value no longer decreases for $5000$ iterations. We use $L_{1}$ loss as our loss function. 

\paragraph{Defining the Metrics.}
We now more thoroughly define all the metrics we used to measure the performance of the CARLA experiments, following standard protocol~\citep{codevilla2018end,codevilla2019exploring,chen2020learning}:
\begin{itemize}[leftmargin=*]
    \item The \%success is the number of episodes fully completed by the ego vehicle among all the 100 predefined test episodes. Higher is better.
    \item \#collision reports the total times the ego car crashes into the pedestrians, vehicles and other obstructions (over 100 test episodes). Lower is better.
    \item For \%progress, we calculate the euclidean distance from the start point to the target point at the beginning of each episode (initial distance) and then compute the distance from the start point to the final point after the episode is over (final distance); and the \%progress is 1 minus the ratio between final distance and initial distance. Higher is better.
    \item The avg.~speed is calculated by dividing the path length that the ego car traveled by the travel time. Higher is better in general, but approaches that drive the vehicle straight into pedestrians at high speed will still get avg.~speed, so this metric is not very informative without the context of the other metrics. 
\end{itemize}

\paragraph{Reporting Results In Terms of All Four Metrics.} 
In Sec 5 in the paper, we only had space to report the quantitative results in terms of \%success, but we remarked on broad trends in terms of the other metrics above too. We now report those complete results in Table~\ref{tab: carla-withspeed} (for unmodified CARLA, including the agent velocity inputs, corresponding to Table 1 in the paper) and Table~\ref{tab: carla-nospeed} (for the CARLA-w/o-speed setting with increased partial observability).

\paragraph{Expanded Fig 6.} Figure 6 in the paper compares the imitation errors of various models on changepoint and non-changepoint samples.
We now report those same results in tabular form for increased precision, in Table~\ref{tab: loss-info}.

\begin{table*}[t]
\caption{The loss of different types of samples. Our method significantly reduces the loss of hard samples, thus reducing the empirical risks among the whole dataset. The standard deviation values are smaller than $1 \times 10^{-3}$ so we don't put them here.}
\label{tab: loss-info}
\vskip 0.15in
\begin{center}
\begin{small}
\begin{sc}
\begin{tabular}{c|c|cc|cc|cc}
\toprule
\multirow{2}{*}{Experiment} & \multirow{2}{*}{Method} & \multicolumn{2}{c|}{unweighted loss$\times 10^{-2}$} & \multicolumn{2}{c|}{changepoint loss$\times 10^{-2}$} &  \multicolumn{2}{c}{copycat sample loss$\times 10^{-2}$} \\
 &  & train & val & train & val & train & val \\
\midrule
  &  BC-SO   &  1.769  &  1.464  &  4.995  &  4.222  &  1.410  &  1.157  \\
w/-speed  &  BC-OH  &  1.549   &  1.395  &  5.721  &  4.808  &  1.086  &  1.016  \\
  &  Ours (step)  &  1.742  &  1.537  &  4.545  &  3.731  & 1.431 &  1.294  \\
\midrule
  &  BC-SO   &  5.350  &  4.989  &  11.213  &  11.059  &  4.710  &  4.316  \\
w/o-speed  &  BC-OH  &  2.172  &  2.055  &  8.194  &  7.588  &  1.504  &  1.441  \\
  &  Ours (step)  &  1.671  &  1.491  &  4.985  &  4.132  &  1.303  &  1.198  \\
\bottomrule
\end{tabular}
\end{sc}
\end{small}
\end{center}
\vskip -0.1in
\end{table*}

\subsection{Additional Details on MuJoCo Experiments}

\paragraph{Architectures and Training Details.}
We use ResNet18 as our backbone and a two-layer MLP with 300 hidden units on top of it to get the predicted action. We use Adam optimizer with initial learning rate $2 \times 10^{-4}$ and reduce the learning rate by a factor of 10 every 100k iterations during training. We use a batchsize of 128. We train the imitation agent for 300k iterations until convergence. We use $L_{2}$ as our loss function.

\subsection{Hyperparameter Choices}
Since there are only few hyperparameters in our method, we can perform grid search to find the best set of hyperparameters based on evaluation reward. For $\rm{softmax}$ weighting function, we select its temperature $\tau$ from \{0.1, 0.2, 0.5, 1, 5, 10\} and for $\rm{step}$ weighting function, we select its threshold $\rm{THR}$ from \{10\%, 20\%\} and its weight $W$ from \{3, 5, 10\}.

\paragraph{Our method}
We use $\rm{step}$ function for CARLA environment. In both of \textit{CARLA} and \textit{CARLA-w/o-speed}, we set $\rm{THR}$ to $10\%$ and $W$ to 5, i.e., upweight the top 10\% of samples by a factor of 5.

In \textit{MuJoCo-Image}, we experiment with both $\rm{step}$ and $\rm{softmax}$ function. For the $\rm{softmax}$ function, we use a temperature of 0.2 for Hopper and Walker2D and 0.1 for HalfCheetah. For the $\rm{step}$ function, we set the $\rm{THR}$ to $10\%$ and $W$ to 5.

\paragraph{Baselines}
In HistoryDropout, we set the dropout rate to 0.5 for all environments.
In FCA, we set different adversarial loss weight for different environment to balance the adversarial loss and the imitation loss. More specifically, the adversarial loss weights for the different environments are \textit{CARLA} and \textit{CARLA-w/o-speed}: 0.2, Hopper: 0.1, HalfCheetah and Walker2D: 0.5.

\subsection{Sensitivity Study}
To study the sensitivity of out method to the hyperparameter choices, we ran some additional experiments by perturbing the hyperparameters. For Walker 2D, we set the softmax temperature to 0.1, 0.2, and 0.5, and the test rewards are $765 \pm 103$, $769 \pm 97$, and $724 \pm 61$. And for CARLA, the weight $W$ of step function is set to 4, 5 and 6, producing success rate 41.89, 43.44, and 41.33\%. 
These results suggest that our method is pretty robust to the hyperparameter choices.

\subsection{Additional Sandbox Environment: ToyCar}

\begin{figure}[t]
    \centering
    \includegraphics[width=0.45\textwidth]{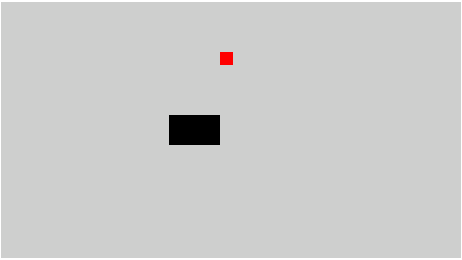}
    \caption{Snapshot of the ToyCar environment.
    }
    \label{fig: toycar}
\end{figure}

In addition to the standard environments in the paper, we constructed a simple sandbox environment for fast experimentation, that we will refer to as ToyCar.
The environment consists of a car on a straight road with a traffic light that stays on red or green for random lengths of time as shown in Fig~\ref{fig: toycar}. The car must successfully drive through the road from left to right without breaking red lights. The car follows point mass double integrator dynamics, and has two actions, throttle, and brake, that apply fixed positive and negative accelerations. There is a fixed upper limit on the velocity. The expert is a rule-based agent and has access to the full observation, i.e. the car position, the velocity, the traffic light position, the traffic light status and the time remaining for the current status. All the imitator policies act on 3x128x128 images.

\begin{table}[h]
\caption{ToyCar results. }
\label{tab: ToyCar-results}
\vskip 0.15in
\begin{center}
\begin{scriptsize}
\begin{sc}
\begin{tabular}{cccc}
\toprule
 & BC-SO &  BC-OH  &  Ours(step)  \\
\midrule
\%Success Rate  &  97.2 $\pm$ 0.7  &  95.1 $\pm$ 4.5  &  97.8 $\pm$ 0.5 \\
\bottomrule
\end{tabular}
\end{sc}
\end{scriptsize}
\end{center}
\vskip -0.1in
\end{table}

\paragraph{Results}
We train the BC-SO, BC-OH and our method with 1K expert samples, and test them in the environment. The test success rates are shown in the Table~\ref{tab: ToyCar-results}. All results are reported over 5 trials. Quantitative results mirror those reported in the paper, but gains are relatively small due to the fact that the environment is very simple. Qualitatively, we observed instances of similar problems with BC-OH to the inertia problem reported before in~\citet{codevilla2019exploring}, namely, the car sometimes stops at a seemingly random location and refuses to start again. Our method successfully removes those failure cases.

\subsection{Low-Dimensional Environments: MuJoCo-State}
Finally, while in the paper, we showed results for image-based MuJoCo settings, we now report results in low-dimensional partially observed MuJoCo settings. 
Following \citet{chuan2020fighting}, we remove the velocity and external force information from the full state to make the environment partially observed.

The environmental rewards are shown in Tab~\ref{tab: mujoco-state-results}. As in the image-based settings, our method successfully outperforms BC-OH easily in all these settings. Further, our results in these environments are comparable to FCA~\cite{chuan2020fighting}. We clearly outperform FCA on Walker2D, perform on par with FCA on HalfCheetah, and perform worse on Hopper. However, the FCA method, reliant on adversarial training, scales poorly to higher-dimensional image-based environments as mentioned in~\citet{chuan2020fighting}, and also verified throughout our other results.

\begin{table}[htb]
\caption{MuJoCo-State results.}
\label{tab: mujoco-state-results}
\vskip 0.15in
\begin{center}
\begin{scriptsize}
\begin{sc}
\begin{tabular}{cccc}
\toprule
 & Hopper &  HalfCheetah  &  Walker2D  \\
\midrule
BC-SO  &  275 $\pm$ 40  &  -38 $\pm$ 36  &  363 $\pm$ 86 \\
BC-OH  &  293 $\pm$ 83  &  820 $\pm$ 60  &  592 $\pm$ 124 \\
FCA  &  \textbf{1086 $\pm$ 262}  &  \textbf{1250 $\pm$ 42}  &  1296 $\pm$ 288 \\
Ours &  641 $\pm$ 7  &  \textbf{1023 $\pm$ 75}  &  \textbf{1460 $\pm$ 169} \\

\bottomrule
\end{tabular}
\end{sc}
\end{scriptsize}
\end{center}
\vskip -0.1in
\end{table}

\subsection{Additional Baseline: Category Frequency Weighting with Discovered Categories}
As we mentioned in Sec 2 in the paper, prior approaches \cite{cui2019class,cao2019learning} have proposed rebalancing data in supervised learning settings based on the frequencies of categories, so that low-frequency categories are upsampled to correct data imbalance issues. We now report the results of a baseline that discovers action categories in an unsupervised manner and apply a similar strategy. This evaluates a different reweighting strategy to the APE-based strategy proposed in the paper.

We assume that there are 6 typical scenarios in urban autonomous driving tasks, i.e. going straight, turning right, turning left, accelerating, slowing down and parking, so we use the k-means algorithm to cluster the actions in the training dataset into $k=6$ clusters as categories. And we use category frequency based weighting function $w_{i}=\frac{\sum_{j=1}^{6}n_{j}}{n_{i}}$ to reweight each sample in the cluster $i$, where $n_j$ is the number of samples in cluster $j$.

This naive reweighting strategy-based baseline does not perform very well. The results are shown in the last rows in Table~\ref{tab: carla-withspeed} and Table~\ref{tab: carla-nospeed}.

\end{document}